\documentclass{article}
\usepackage[preprint]{neurips_2022}
\usepackage[utf8]{inputenc} 
\usepackage[T1]{fontenc}    
\usepackage{hyperref}       
\usepackage{url}            
\usepackage{booktabs}       
\usepackage{amsfonts}       
\usepackage{nicefrac}       
\usepackage{microtype}      
\usepackage{xcolor}         
\usepackage{graphicx}

\usepackage{caption}
\usepackage{subcaption}
\usepackage{wrapfig}
\usepackage{multirow}
\usepackage{amsmath}
\usepackage{amsfonts}
\usepackage{bm}
\usepackage{amssymb}
\usepackage{amsthm}
\usepackage{verbatim}
\usepackage{longtable}
\usepackage{arydshln}
\usepackage{algorithm}
\usepackage{algpseudocode}
\usepackage{mathtools, nccmath}

\newcommand{\ours}{\texttt{BiT}}

\DeclareMathOperator*{\argmin}{arg\,min}
\DeclarePairedDelimiter{\nint}\lfloor\rceil

\newcommand{\dataset}{\ensuremath{\mathcal{D}}}
\newcommand{\trainDataset}{\ensuremath{\dataset_{\text{train}}}}
\newcommand{\devDataset}{\ensuremath{\dataset_{\text{dev}}}}
\newcommand{\model}{\ensuremath{h}}
\newcommand*\samethanks[1][\value{footnote}]{\footnotemark[#1]}

\title{\ours: Robustly Binarized Multi-distilled Transformer}

\author{
  Zechun Liu\thanks{\hspace{.06in}Equal contribution} \\
  Reality Labs, Meta Inc. \\
  \texttt{zechunliu@fb.com} \\
  \And
  Barlas Oğuz\samethanks \\
  Meta AI \\
  \texttt{barlaso@fb.com} \\
  \And
  Aasish Pappu \\
  Meta AI \\
  \texttt{aasish@fb.com} \\
  \AND
  Lin Xiao \\
  Meta AI \\
  \And
  Scott Yih \\
  Meta AI \\
  \And
  Meng Li \\
  Peking University \\
  \AND
  Raghuraman Krishnamoorthi \\
  Reality Labs, Meta Inc. \\
  \And
  Yashar Mehdad \\
  Meta AI \\
}

\begin{document}
\vspace{-1cm}
\maketitle

\begin{abstract}
Modern pre-trained transformers have rapidly advanced the state-of-the-art in machine learning, but have also grown in parameters and computational complexity, making them increasingly difficult to deploy in resource-constrained environments.  Binarization of the weights and activations of the network can significantly alleviate these issues, however, is technically challenging from an optimization perspective. In this work, we identify a series of improvements that enables binary transformers at a much higher accuracy than what was possible previously. These include a two-set binarization scheme, a novel elastic binary activation function with learned parameters, and a method to quantize a network to its limit by successively distilling higher precision models into lower precision students. These approaches allow for the first time, fully binarized transformer models that are at a practical level of accuracy, approaching a full-precision BERT baseline on the GLUE language understanding benchmark within as little as 5.9\%. Code and models are available at: \url{https://github.com/facebookresearch/bit}.
\end{abstract}
\vspace{-0.6cm}
\section{Introduction}
The past few years have witnessed tremendous advances in almost all applied fields of AI.  It would hardly be a simplification to say that the bulk of these advances was achieved by scaling the transformer architecture \citep{vaswani2017attention} to ever larger sizes with increasing computation budget \citep{devlin2019bert, liu2019roberta, radford2018improving, radford2019language, raffel2020exploring, brown2020language}.  On the other hand, mobile devices and wearables have proliferated and shrunk in size, with stringent requirements for storage, computation and energy consumption.  Consumers demand more portability, while having access to all that current AI technology has to offer.  As a result, the gap between what is possible in AI, and what is deployable has never been wider.

While there is a variety of methods to increase inference efficiency in neural networks (e.g. knowledge distillation, pruning), quantization has some attractive properties and has been widely successful in practice \citep{gholami2021survey}.  For one, storage and latency gains from quantization are deterministically defined for a given quantization level.  For instance, reducing the precision of model parameters by a given factor immediately translates to an identical reduction in storage cost.  Similarly, reducing the precision of arithmetic operations results in a corresponding reduction in computational cost.  Uniform quantization is hardware friendly, making it relatively simple to realize theoretical improvements in practice.

Binarization represents the extreme limit of quantization, promising a 32$\times$ reduction in storage over full-precision (32-bit) models.  Moreover, binary arithmetic completely eliminates multiplications in favor of bit-wise XNOR operations \citep{courbariaux2016binarized, rastegari2016xnor}, enabling even further improvements when using special purpose hardware. Energy efficiency improvements between 100-1000x have been claimed to be possible with binary neural networks (BNNs), over their full-precision counterparts \citep{nurvitadhi2016accelerating}.

The obvious challenge with binarization is the difficulty of optimization.  While all quantization is discontinuous, higher precisions allow approximating the full-precision network to a better extent, where with BNNs, this becomes much harder.  Surprisingly, researchers in computer vision have been able to demonstrate BNNs with remarkable accuracy \citep{liu2018bi, IR-Net, martinez2020training}.  Unfortunately, while these works have mostly been developed on convolutional architectures for image tasks, they have not generalized well to transformer models.  For instance, recent work \citep{qin2021bibert} has shown that a BERT \citep{devlin2019bert} model with binary weights and activations lags its full-precision counterpart by as much as 20 points on average on GLUE dataset. Even for weights-only binarization, the loss landscape was shown to be too irregular, and recent work resorted to complex and specialized methods such as weight-splitting from half-width models to achieve a reasonable accuracy \citep{bai2021binarybert}.

With this background, we tackle the problem of fully binarizing transformer models to a high level of accuracy.  With the expansion trend of transformers towards becoming the standard architecture choice for all fields of AI, we believe a solution to this problem could be highly impactful.

Our approach follows the same paradigm as previous work, based on knowledge distillation \citep{hinton2015distilling} from higher precision models using the straight-through estimator (STE) of \citet{bengio2013estimating}.  In view of the optimization difficulties, we take the following steps to ensure that the student and teacher models are well-matched:
\begin{itemize}
    \item  In Section \ref{sec:baseline} we describe a robust binarization framework, which allows the binary student network to better match the output distribution of the teacher.  This allows us to achieve SoTA results for extreme activation quantization with BERT, producing models with little loss in accuracy down to a quantization level of binary weights and 2-bit activations and improves over previous setups by large margins in the fully binary (1-bit) setting.  It also leads to competitive results for weight binarization with 4-bit activations using a single knowledge distillation step.
    \item  To further improve binary models, we propose a multi-distillation approach, described in Section \ref{sec:whisky}.  Instead of distilling directly into a fully binary model, we first distill an intermediate model of medium precision and acceptable accuracy.  This model then becomes the teacher in the next round of distillation into increasingly quantized models.  Such a method ensures that the student model doesn't drift too far from the teacher, while also ensuring as good an initialization as possible.  We call the resulting model \ours{} \footnote{Short for Binarized Transformer.}.
\end{itemize}

In the vanilla setting without data augmentation, our approach reduces the accuracy gap to a full-precision BERT-base model by half on the GLUE \citep{wang2018glue} benchmark compared to the previous SoTA.  When using data augmentation, we are able to reduce the absolute accuracy gap to only 5.9 points (from over 15 points previously).  In addition to the fully binary setting, we also report SoTA results with binary weights and 2-bit activations, where our models trail the full-precision baseline by only 3.5 points.

\section{Background}
\subsection{Transformer architecture}
The transformer model of \citet{vaswani2017attention} is composed of an embedding layer, followed by $N$ transformer blocks and a linear output layer.  Each transformer block consists of a multi-head attention layer followed by a feed-forward network, as shown in Figure~\ref{fig:overview}.  The multi-head attention layer is a concatenation of $K$ scaled dot-product attention heads, defined by:
\[ \text{Attention}(Q, K, V) = \text{softmax}\left(\frac{QK^T}{\sqrt{d_k}}\right) V \]
where $d_k$ is the dimension of each key, $Q, K, V$ are weight matrices for the query, key and value respectively.  As such, the computation in a transformer model is limited to linear matrix multiplications and additions, pointwise non-linearities (most commonly Sigmoid~\citep{han1995influence}, GeLU~\citep{hendrycks2016gaussian} or ReLU~\citep{nair2010rectified} ) and the Softmax operation~\citep{bridle1989training}.

\subsection{Quantization}
A vector $w$ is uniformly quantized to $b$-bit precision, if its entries are restricted to the set $\{0, 1, \ldots, 2^b-1\}$ for asymmetric case or $\{-2^b, -2^b+1, \ldots, 2^b-1\}$ for symmetric case, up to a real-valued scale $\alpha$. This allows vector operations to utilize lower precision arithmetic, making them more efficient by a factor of $\frac{B}{b}$ compared to full-precision calculation using $B$ bits.  The scaling operation is still in higher precision, but if the dimensionality of $w \gg \frac{B}{b}$, then the extra computation is negligible.

A neural network with parameters quantized to $b_w$ bits takes up $\frac{B}{b_w}$ times less space. However, to take advantage of lower-precision arithmetic, the input vectors (activations) to each vector/matrix operation also need to be quantized.  A network which has weights quantized to $b_w$ bits and activations quantized to $b_a$ bits is denoted as $\text{W}b_w\text{A}b_a$.  In this work, we're specifically interested in W1A1 transformers.  Binary arithmetic is especially attractive, since multiplications reduce to \text{XNOR} operations, and can be implemented orders of magnitude more efficiently using specialized hardware \citep{nurvitadhi2016accelerating}.

\subsection{Knowledge distillation}
Knowledge distillation (KD) \citep{hinton2015distilling} is a technique whereby a student network can be trained to mimic the behavior of a teacher network.  This is especially useful when the student network is more efficient and easier to deploy than the more complex and cumbersome teacher.  The basic way of performing KD is by using the output distribution of the teacher model ($\mathbf{p}$) as soft targets for training the student model.  If $\mathbf{q}$ is the student model's output, then we have the loss term:
\begin{equation}
    \mathcal{L}_{\text{logits}} = \text{KL}(\mathbf{p}, \mathbf{q})
\end{equation}
The advantage of KD over simple supervised training of a more efficient model is that the teacher model provides a richer training signal including model confidence for each output class.

For computer vision tasks, distilling the final logits solely works well for binary neural networks~\citep{liu2020reactnet}.
If the student and teacher architectures are compatible, one can also distill intermediate activations for faster convergence and better transfer and generalization \citep{aguilar2020knowledge}:
\begin{equation}
    \mathcal{L}_{\text{reps}} = \sum_i ||r_i^s - r_i^t||^2,
\end{equation}
where $r_i^s$ and $r_i^t$ are the corresponding transformer block output activations from student and teacher.

\section{Robust binarization setup}
\label{sec:baseline}
In this section we first bring together some best practices and minor improvements which we have found helpful in simplifying previous work and building a strong baseline.  Then we present a novel activation binarization scheme, which we will show to be critical to achieve good performance.

\subsection{Two-set binarization scheme}
\label{sec:two_set}

In contrast to convolutional neural networks on images where activations exhibit comparable distributions, different activations in transformer blocks are performing different functionalities, and thus vary in their output distributions. In particular, these activations can be divided into two categories: the activations after Softmax/ReLU layer that contains positive values only and the remaining activations with both positive and negative values
(e.g., after matrix multiplication).
If we denote by $\mathbf{X_R}$ the vector of activation values, then the two cases are $\mathbf{X}_\mathbf{R}^i \in \mathbb{R}_+$ and $\mathbf{X}_\mathbf{R}^i \in \mathbb{R}$ respectively.

\begin{figure}[t!]
    \centering
    \includegraphics[width=0.5\linewidth]{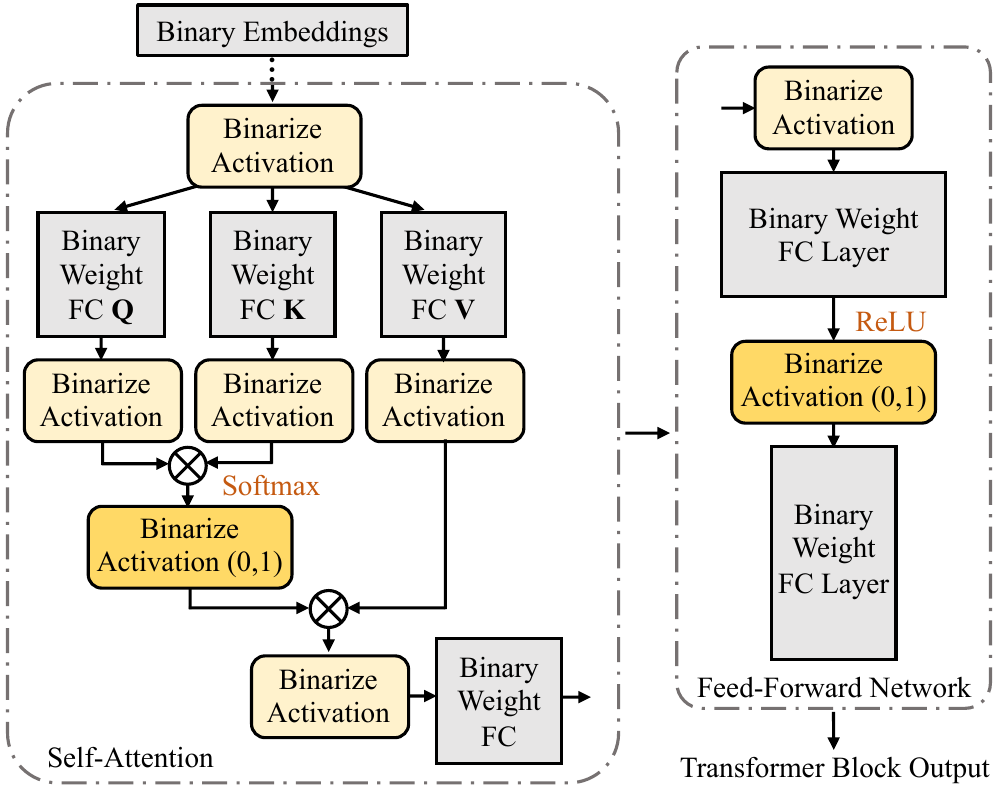}
    \caption{Overview of \ours{}. A transformer block contains the multi-head self-attention and feed-forward network. We binarize all the weights to \{-1, 1\} in the Embedding/Fully-Connected layers and binarize activations to \{0, 1\} for ReLU/Softmax outputs and to \{-1, 1\} for other layers.}
    \label{fig:overview}
\end{figure}

For the former set, mapping to the binary levels $\{-1, 1\}$ would result in a severe distribution mismatch.  Therefore we instead map non-negative activation layers to $\mathbf{\hat{X}_B} \in \{0, 1\}^n$ and binarize activation layers with $\mathbf{X_R} \in \mathbb{R}^n$ to $\mathbf{\hat{X}_B} \in \{-1, 1\}^n$, shown in Figure~\ref{fig:overview}. A prior work BiBERT~\citep{qin2021bibert} also suggests binarizing attention to $\{0, 1\}$, but with $\rm{bool}$ function replacing $\rm{SoftMax}$, while we empirically find that simply binarizing attentions after $\rm{SoftMax}$ to $\{0, 1\}$ works better and binarizing $\rm{ReLU}$ output to $\{0, 1\}$ instead of $\{-1, 1\}$ brings further improvements. (See Section~\ref{sec:vs_biattention} for details).

\textbf{Optimal scaling factor in two sets}
Additionally, we apply a layer-wise scaling factor to binarized activations to reduce the binarization error, \textit{i.e.}, $\mathbf{X_B} = \alpha \mathbf{\hat{X}_B}$. The optimal values of $\alpha$ are different for the $\mathbf{\hat{X}_B} \in \{0, 1\}^n$ and $\mathbf{\hat{X}_B} \in \{-1, 1\}^n$ cases and can be calculated by minimizing the $l2$ error:
\begin{equation}
\label{eq:l2_err}
    \begin{split}
    & \mathcal{J}(\alpha) = ||\mathbf{X_R} - \alpha\mathbf{\hat{X}_B}||^2 \\
    & \alpha^* = \argmin_{\alpha \in \mathbb{R}_+} \mathcal{J} ( \alpha)
    \end{split}
\end{equation}
Following XNOR-Net~\citep{rastegari2016xnor}, by expanding Eq.~\ref{eq:l2_err}, we have
\begin{equation}
\mathcal{J}(\alpha) = \alpha^2\hat{\mathbf{X}_B}^T\mathbf{\hat{X}_B} - 2\alpha\mathbf{X_R}^T\mathbf{\hat{X}_B} + \mathbf{X_R}^T\mathbf{X_R}
\end{equation}
For the layers with $\mathbf{X_R} \in \mathbb{R}^n$ we follow the traditional methods
of binarizing activations
\citep{rastegari2016xnor,liu2018bi}
by taking the sign of real-valued activations:
\begin{equation}
\mathbf{\hat{X}_B}^i = {\rm Sign}(\mathbf{X}_\mathbf{R}^i) = \left\{
         \begin{array}{lr}
         \vspace{0.3em}
         - 1, & {\rm if} \ \mathbf{X}_\mathbf{R}^i < 0 \\
         + 1, & {\rm if} \ \mathbf{X}_\mathbf{R}^i \geqslant 0
         \end{array}
         \right.
\end{equation}
In that case, $\hat{\mathbf{X}_B}^T\mathbf{\hat{X}_B}=n_{\mathbf{X_R}}$, where $n_{\mathbf{X_R}}$ is number of elements in $\mathbf{X_R}$, and $\alpha^*$ can be solved as:
\begin{equation}
\alpha^* =  \frac{\mathbf{X_R}^T\mathbf{\hat{X}_B}}{n_{\mathbf{X_R}}} = \frac{||\mathbf{X_R}||_{l1}}{n_{\mathbf{X_R}}}
\end{equation}
For the activations in attention layers or after the ReLU non-linearity layers with $\mathbf{X_R} \in \mathbb{R}_+^n$, we binarize the activations to $\mathbf{\hat{X}_B} \in \{0, 1\}^n$ by rounding the real-valued activations:
\begin{equation}
\mathbf{\hat{X}_B}^i = \nint{{\rm Clip}(\mathbf{X}_\mathbf{R}^i, 0, 1)} = \left\{
         \begin{array}{lr}
         \vspace{0.3em}
         0, & {\rm if} \ \mathbf{X}_\mathbf{R}^i < 0.5 \\
         1, & {\rm if} \ \mathbf{X}_\mathbf{R}^i \geqslant 0.5
         \end{array}
         \right.
\end{equation}
In that case, $\hat{\mathbf{X}_B}^T\mathbf{\hat{X}_B}=n_{\{\mathbf{X_R} \geqslant 0.5\}}$
where $n_{\{\mathbf{X_R} \geqslant 0.5\}}$ denotes the number of elements in $\mathbf{X_R}$ that are greater than or equal to $0.5$. Then $\alpha^*$ can be solved as:
\begin{equation}
\alpha^* = \frac{||\mathbf{X_R} \cdot \textbf{1}_{\{\mathbf{X_R} \geqslant 0.5\}}||_{l1}}{n_{\{\mathbf{X_R} \geqslant 0.5\}}}
\end{equation}

\subsection{Best practices}\label{sec:minor_improvements}
We performed thorough experimentation and discovered the following modifications to be useful.

\textbf{Simplified knowledge distillation}
Compared to previous BERT model binarization works~\citep{bai2021binarybert,qin2021bibert} which also attempt to distill the attention scores, we provide analysis and experimental results (Section \ref{sec:ablations}) showing that using only $\mathcal{L}_{\text{reps}}$ from transformer block outputs and $\mathcal{L}_{\text{logits}}$ is more effective while being simpler.  We also forego the two-step distillation scheme of~\citet{bai2021binarybert} in favor of a single step, joint distillation, where our training loss is simply $\mathcal{L}_{\text{logits}} + \mathcal{L}_{\text{reps}}$.

\textbf{Mean subtraction in weight binarization} For weight binarization, centeralizing the real-valued weights to be zero-mean before binarization can increase the information carrying capacity of the binary weights. Thus, for weight binarization, we have: $\mathbf{W}_\mathbf{B}^i = \frac{||\mathbf{W}_\mathbf{R}||_{l1}}{n_{\mathbf{W}_\mathbf{R}}}{\rm Sign}(\mathbf{W}_\mathbf{R}^i - \mathbf{\overline{W}_R})$.

\textbf{Gradient clipping} Clipping gradients to $0$ when $\mathbf{X}_\mathbf{R}^i \notin [-1,1]$ (or $\mathbf{X}_\mathbf{R}^i \notin [0,1]$ if $\mathbf{\hat{X}_B} \in \{0, 1\}^n$) is a common technique for training binarized neural networks.  However, we find that clipping weight gradients is harmful for optimization.  Once a weight is outside of the clip range, the gradient is fixed to $0$, preventing further learning.  This is not so for activations, since the activation value changes for each input. As a result, we apply gradient clipping only to activations but not to weights.

\textbf{Non-linearity} We prefer ReLU activations whenever the output range is non-negative.

Combining these, we are able to build a strong baseline, which improves the accuracy by 9.6\% over naively binarized transformers. Additionally, these techniques allow us to train a weight binarized transformer network in a single training step using knowledge distillation (\textit{i.e.}, without resorting to weight splitting as in BinaryBert~\citep{bai2021binarybert}) (See Section \ref{sec:ablations} for details).

\subsection{Elastic binarization function}
\label{sec:activation_scaling}
The fixed scaling and threshold derived previously works reasonably well, but might not be optimal since it ignores the distribution of the variable which is being binarized.  Ideally, these parameters can be learned during training to minimize the target loss.

When using classical binarization methods, \textit{i.e.}, $\mathbf{\hat{X}_B}^i = {\rm Sign}(\mathbf{X}_\mathbf{R}^i)$, the binary output is independent of the scale of the real-valued input. However, in our case where $\mathbf{\hat{X}_B}^i = \nint{{\rm Clip}(\mathbf{X}_\mathbf{R}^i, 0, 1)}$, this independence no longer holds.  Learning the scaling and threshold parameters, and how to approximate the gradients precisely in the process becomes crucial for the final accuracy.

To handle this, inspired by the learnable threshold in ReActNet~\citep{liu2020reactnet}, we propose the elastic binarization function to learn both the scale $ \alpha \in \mathbb{R}_+$ and the threshold $\beta \in \mathbb{R}$:
\begin{equation}
    \mathbf{X}_\mathbf{B}^i = \alpha \mathbf{\hat{X}_B}^i = \alpha \nint{{\rm Clip}(\frac{\mathbf{X}_\mathbf{R}^i - \beta}{\alpha}, 0, 1)}
\end{equation}
In the function, we initialize $\alpha$ with $\alpha^*$ in Sec.~\ref{sec:two_set} and $\beta$ to be $0$, and train it with gradients from the final loss. To back-propagate the gradients to $\alpha$ through the discretized binarization function, we follow the practice in ~\citet{choi2018pact, zhou2016dorefa,esser2019learned} to use straight-through estimator (STE)~\citep{bengio2013estimating} to bypass the incoming gradients to the round function to be the outgoing gradients:
\begin{equation}
    \begin{split}
    \frac{\partial\mathbf{X}_\mathbf{B}^i}{\partial \alpha}
    & = \mathbf{\hat{X}_B}^i + \alpha \frac{\partial\mathbf{\hat{X}_B}^i}{\partial \alpha}  \\
    & \!\!\! \overset{STE}{\approx} \mathbf{\hat{X}_B}^i + \alpha \frac{\partial{\rm Clip}(\frac{\mathbf{X}_\mathbf{R}^i - \beta}{\alpha}, 0, 1)}{\partial \alpha} \\
    & =  \left\{
         \begin{array}{lr}
         0,  &{\rm if} \ \ \mathbf{X}_\mathbf{R}^i < \beta \\
         \frac{\beta - \mathbf{X}_\mathbf{R}^i}{\alpha},  & {\rm if} \ \  \beta \leqslant \mathbf{X}_\mathbf{R}^i < \alpha / 2 + \beta \\
         1 - \frac{\mathbf{X}_\mathbf{R}^i - \beta}{\alpha},  \ \ \ \ & {\rm if} \ \ \alpha / 2 + \beta \leqslant \mathbf{X}_\mathbf{R}^i < \alpha + \beta \\
         1,  & {\rm if} \ \ \mathbf{X}_\mathbf{R}^i \geqslant \alpha + \beta \\
         \end{array}
         \right.
    \end{split}
\end{equation}
Then the gradients \textit{w.r.t.} $\beta$ can be similarly calculates as:
\begin{equation}
\frac{\partial\mathbf{X}_\mathbf{B}^i}{\partial \beta} \overset{STE}{\approx} \alpha \frac{\partial{\rm Clip}(\frac{\mathbf{X}_\mathbf{R}^i - \beta}{\alpha}, 0, 1)}{\partial \beta}  = \left\{
         \begin{array}{lr}
         \vspace{0.2em}
         - 1, & {\rm if} \ \beta \leqslant \mathbf{X}_\mathbf{R}^i < \alpha + \beta \\
         0, & {\rm otherwise}
         \end{array}
         \right.
\end{equation}

For the layers that contain both positive and negative real-valued activations \textit{i.e.},  $\mathbf{X_R} \in \mathbb{R}^n$, the binarized values $\mathbf{\hat{X}_B} \in \{-1, 1\}^n$ are indifferent to the scale inside the Sign function: $ \mathbf{X}_\mathbf{B}^i = \alpha \cdot {\rm Sign}(\frac{\mathbf{X}_\mathbf{R}^i - \beta}{\alpha}) = \alpha \cdot {\rm Sign}(\mathbf{X}_\mathbf{R}^i - \beta)$.
In that case, since the effect of scaling factor $\alpha$ inside the Sign function can be ignored, the gradient \textit{w.r.t.} $\alpha$ can be simply calculated as $\frac{\partial\mathbf{X}_\mathbf{B}^i}{\partial \alpha} = {\rm Sign}(\mathbf{X}_\mathbf{R}^i - \beta)$.

In our ablations (Section \ref{sec:ablations} and ~\ref{sec:vs_reactnet}) we show that using this simple elastic binarization function can bring a 15.7\% accuracy boost over our strong baseline on the GLUE benchmark.

\section{Multi-distilled binary transformer}\label{sec:whisky}
\begin{algorithm}[t]
\caption{\ours: Multi-distillation algorithm}
\label{alg:algorithm}
\small
\begin{algorithmic}[1]
\Require $\trainDataset, \devDataset$ \Comment{Training Data}
\Require $\model_0$ \Comment{Full-precision Model}
\Require $\mathbf{Q} = \{(b_w^1, b_a^1), \ldots, (b_w^k, b_a^k)\}$ \Comment{Quantization Schedule}
\State $\model_{\text{teacher}} \gets \ \model_0$
\For{$b_w^i, b_a^i$ in $\mathbf{Q}$}
\State $\model_{\text{student}} \gets \ \text{Quantize}(\model_{\text{teacher}}, b_w^i, b_a^i)$
\State $\text{KnowledgeDistill}(\model_{\text{student}}, \model_{\text{teacher}}, \trainDataset, \devDataset)$
\State $\model_{\text{teacher}} \gets \ \model_{\text{student}}$
\EndFor
\State \Return $\model_{\text{student}}$
\end{algorithmic}
\end{algorithm}

Classical knowledge distillation (KD) \citep{hinton2015distilling} trains the outputs (\textit{i.e.}, logits) of a student network to be close to those of a teacher, which is typically larger and more complex.  This approach is quite general, and can work with any student-teacher pair which conforms to the same output space.  However, in practice, knowledge transfer happens faster and more effectively if the intermediate representations are also distilled \citep{aguilar2020knowledge}.  This approach has been found useful when distilling to student models with similar architecture \citep{sanh2019distilbert}, and in particular for quantization~\citep{bai2021binarybert,kim2019qkd}.

Note that having a similar student-teacher pair is a requirement for distilling representations.  While how similar they need to be is an open question, intuitively a teacher which is architecturally closer to the student should make transfer of internal representations easier.  In the context of quantization, it is easy to see that lower precision students are progressively less similar to the full-precision teacher, which is one reason why binarization is difficult.

This suggests a multi-step approach, where instead of directly distilling from a full-precision teacher to the desired quantization level, we first distill into a model with sufficient precision in order to preserve quality.  This model can then be used as a teacher to distill into a further quantized student.  This process can be repeated multiple times, while at each step ensuring that the teacher and student models are sufficiently similar, and the performance loss is limited.  This multi-distillation approach is sketched in Algorithm \ref{alg:algorithm}.

The multi-step distillation follows a \emph{quantization schedule}, $\mathbf{Q} = \{(b_w^{\ 1}, b_a^{\ 1}), (b_w^{\ 2}, b_a^{\ 2}), \ldots, (b_w^{\ k}, b_a^{\ k})\}$ with $(b_w^{\ 1}, b_a^{\ 1}) > (b_w^{\ 2}, b_a^{\ 2}) > \ldots > (b_w^{\ k}, b_a^{\ k})$\footnote{$(a,b) > (c,d)$ if $a > c$ and $b \ge d$ or $a \ge c$ and $b > d$.}.  $(b_w^{\ k}, b_a^{\ k})$ is the target quantization level, which is in our case binary for both weights and activations.  In practice, we find that down to a quantization level of W1A2, we can distill models of reasonable accuracy in single shot, following the best practices outlined in Section \ref{sec:minor_improvements} (See our 1-1-2 baseline results in Table \ref{tab:glue}).  As a result, we follow a fixed quantization schedule, W32A32 $\rightarrow$ W1A2 $\rightarrow$ W1A1.  This is not necessarily optimal, and how to efficiently find the best quantization schedule is an interesting open problem.  We present our initial explorations towards this direction in Section \ref{sec:paths}.

Combining the elastic binary activations with multi-distillation we obtain \ours{}, the robustly binarized multi-distilled transformer.  Note that \ours{} simultaneously ensures good initialization for the eventual (in our case binary) student model.  Since the binary loss landscape is highly irregular, good initialization is critical to aid optimization.  Previous work has proposed progressive distillation \citep{zhuang2018towards, yang2019synetgy} to tackle this problem, wherein the student network is quantized at increasing severity as the training progresses.  However, this method does not prevent the student network from drifting away from the teacher, which is always the full-precision model.  We compare to progressive distillation in Section~\ref{sec:progressive_ablation}.

\section{Main results}\label{sec:main_results}
We follow recent work \citep{bai2021binarybert, qin2021bibert} in adopting the experimental setting of \citet{devlin2019bert}, and use the pre-trained BERT-base as our full-precision baseline.  We evaluate on GLUE~\citep{wang2018glue}, a varied set of language understanding tasks (see Section~\ref{sec:glue} for a full list), as well as SQuAD (v1.1) \citep{rajpurkar2016squad}, a popular machine reading comprehension dataset.

\subsection{GLUE results}
\begin{table}[t]
\renewcommand\arraystretch{0.6}
\centering
\caption{Comparison of BERT quantization methods on the GLUE dev set. The E-W-A notation refers to the quantization level of embeddings, weights and activations. $\ddag$ denotes distilling binary models using full-precision teacher without using multi-distill technique in Section \ref{sec:whisky}. *Data augmentation is not needed for MNLI, QNLI, therefore results in the data augmentation section are identical to that without data augmentation for these datasets.}
\label{tab:glue}
\setlength{\tabcolsep}{1.4mm}
{\fontsize{6.5pt}{\baselineskip}\selectfont
\begin{tabular}{lcccccccccccc}
\hline
\vspace{0.25pt}
\textbf{Quant} & \begin{tabular}[c]{@{}c@{}}\textbf{\#Bits}\end{tabular} & \begin{tabular}[c]{@{}c@{}}\textbf{Size}$_\text{ (MB)}$\end{tabular} & \begin{tabular}[c]{@{}c@{}}\textbf{FLOPs}$_\text{ (G)}$\end{tabular} & \begin{tabular}[c]{@{}c@{}}\textbf{MNLI}$_\text{-m/mm}$\end{tabular} & \textbf{QQP} & \textbf{QNLI} & \textbf{SST-2} & \textbf{CoLA} & \textbf{STS-B} & \textbf{MRPC} & \textbf{RTE} & \textbf{Avg.} \\ 
\hline
BERT & 32-32-32 & 418 & 22.5 & 84.9/85.5 & 91.4 & 92.1 & 93.2 & 59.7 & 90.1 & 86.3 & 72.2  & 83.9 \\ 
\hdashline[0.8pt/1pt]
\multicolumn{3}{l}{\textit{Without data augmentation}} &&&&&&&&&& \\
Q-BERT & 2-8-8 & 43.0 & 6.5 & 76.6/77.0 & -- & -- & 84.6 & -- & -- & 68.3 & 52.7 & -- \\
Q2BERT & 2-8-8 & 43.0 & 6.5 & 47.2/47.3 & 67.0 & 61.3 & 80.6 & 0 & 4.4 & 68.4 & 52.7 &  47.7 \\
{TernaryBERT} & 2-2-8 & 28.0 & 6.4 & 83.3/83.3 & 90.1 & -- & -- & 50.7 & -- & 87.5 & 68.2 & -- \\
{BinaryBERT} & 1-1-8 & 16.5 & 3.1 & 84.2/84.7 & 91.2 & 91.5 & 92.6 & 53.4 & 88.6 & 85.5 & 72.2 & 82.7 \\
{BinaryBERT} & 1-1-4 & 16.5 & 1.5 & 83.9/84.2 & 91.2 & 90.9 & 92.3 & 44.4 & 87.2 & 83.3 & 65.3 & 79.9 \\
{BinaryBERT} & 1-1-2 & 16.5 & 0.8 & 62.7/63.9 & 79.9 & 52.6 & 82.5 & 14.6 & 6.5 & 68.3 & 52.7 &  53.7 \\
{BinaryBERT} & 1-1-1 & 16.5 & 0.4 & 35.6/35.3 & 66.2 & 51.5 & 53.2 & 0 & 6.1 & 68.3 & 52.7 & 41.0 \\
{BiBERT} & 1-1-1 & 13.4 & 0.4 & 66.1/67.5 & 84.8 & 72.6 & 88.7 & 25.4 & 33.6 & 72.5 & 57.4 & 63.2 \\
\hdashline[0.8pt/1pt]
\ours{} $\ddag$ & 1-1-4 & 13.4 & {1.5} & 83.6/84.4 & 87.8 & 91.3 & 91.5 & 42.0 & 86.3 & 86.8 & 66.4 & 79.5 \\
\ours{} $\ddag$ & 1-1-2 & 13.4 & {0.8} & 82.1/82.5 & 87.1 & 89.3 & 90.8 & 32.1 & 82.2 & 78.4 & 58.1 & 75.0\\
\ours{} $\ddag$ & 1-1-1 & 13.4 & {0.4} & 77.1/77.5 & 82.9 & 85.7 & 87.7 & 25.1 & 71.1 & 79.7 & 58.8 & 71.0\\ 
\ours{} & \textbf{1-1-1} & \textbf{13.4} & \textbf{0.4} & \textbf{79.5}/\textbf{79.4} & \textbf{85.4} & \textbf{86.4} & \textbf{89.9} & \textbf{32.9} & \textbf{72.0} & \textbf{79.9} & \textbf{62.1} & \textbf{73.5} \\
\hline
\multicolumn{3}{l}{\textit{With data augmentation}} &&&&&&&&&& \\
{TernaryBERT} & 2-2-8 &  28.0 & 6.4 & 83.3/83.3* & 90.1* & 90.0 & 92.9 & 47.8 & 84.3 & 82.6 & 68.4 & 80.3\\
{BinaryBERT} & 1-1-8 & 16.5 & 3.1 & 84.2/84.7* & 91.2* & 91.6 & 93.2 & 55.5 & 89.2 & 86.0 & 74.0 & 83.3 \\
{BinaryBERT} & 1-1-4 & 16.5 & 1.5 & 83.9/84.2* & 91.2* & 91.4 & 93.7 & 53.3 & 88.6 & 86.0 & 71.5 & 82.6 \\
{BinaryBERT} & 1-1-2 & 16.5 & {0.8}  & 62.7/63.9* & 79.9* & 51.0 & 89.6 & 33.0 & 11.4 & 71.0 & 55.9 & 57.6 \\
{BinaryBERT} & 1-1-1 & 16.5 & {0.4}  & 35.6/35.3* & 66.2* & 66.1 & 78.3 & 7.3 & 22.1 & 69.3 & 57.7 & 48.7 \\
{BiBERT} & 1-1-1 & 13.4 & 0.4 & 66.1/67.5* & 84.8* & 76.0 & 90.9 & 37.8 & 56.7 & 78.8 & 61.0 & 68.8 \\
\hdashline[0.8pt/1pt]
\ours{} $\ddag$ & 1-1-2 & 13.4 & {0.8}  & 82.1/82.5* & 87.1* &  88.8 & 92.5 & 43.2 & 86.3 & 90.4 & 72.9 & 80.4 \\
\ours{} $\ddag$ & 1-1-1 & 13.4 & {0.4}  & 77.1/77.5* & 82.9* &  85.0 & 91.5 & 32.0 & 84.1 & 88.0 & 67.5 & 76.0 \\
\ours{} & \textbf{1-1-1} & \textbf{13.4} & \textbf{0.4} & \textbf{79.5}/\textbf{79.4}* & \textbf{85.4}* & \textbf{86.5} & \textbf{92.3} & \textbf{38.2} & \textbf{84.2} & \textbf{88.0} & \textbf{69.7} & \textbf{78.0} \\
\hline
\end{tabular}}
\end{table}

Our main results on the GLUE benchmarks are presented in Table \ref{tab:glue}.  In the setting without data augmentation, where we only use the original training samples for knowledge distillation, we are able to reduce the gap to the full precision baseline by 49.8\%, \textit{i.e.}, from 20.7 in~\citep{qin2021bibert} to 10.4 points.  We also see that our baseline models with elastic activation binarization already improve previous SoTA by large margins.

In the binary weight setting (4-bit activations), we can match or outperform \citet{bai2021binarybert} without the need for pre-training half-width models and subsequently splitting weights.  This result should make binary weight models much easier to implement and deploy.

We also set a new state of the art for binary weight 2-bit activation (W1A2) models, with only a 3.5 point degradation compared to the full-precision baseline (using data augmentation).  While not as efficient as binary, 2-bit arithmetic can also be performed without multiplications, making it a good efficient alternative in applications where the performance cost of going to fully binary is significant.

\subsubsection{Data augmentation}
From Table \ref{tab:glue}, it can be observed that the datasets with small training sets still have a large gap from the full-precision baseline.  As a result, we employ data augmentation heuristics (following the exact setup in~\citet{TernaryBERT}) on the datasets with small training sets (all except MNLI, QNLI) to take better advantage of our model's strong representational capability. This further reduces the quantization gap, with our models eventually trailing the full-precision model by only 5.9 points on average on the GLUE benchmark.

\begin{table}[t]
\renewcommand\arraystretch{0.6}
\centering
\caption{Ablation study on the effects of each component on GLUE dataset without data augmentation.}
\label{tab:ablation_glue}
\setlength{\tabcolsep}{1.4mm}
{\fontsize{6.5pt}{\baselineskip}\selectfont
\begin{tabular}{llcccccccccccc}
\hline
\vspace{0.25pt}
& \textbf{Quant} & \begin{tabular}[c]{@{}c@{}}\textbf{MNLI}$_\text{-m/mm}$\end{tabular} & \textbf{QQP} & \textbf{QNLI} & \textbf{SST-2} & \textbf{CoLA} & \textbf{STS-B} & \textbf{MRPC} & \textbf{RTE} & \textbf{Avg.} \\ 
\hline
1 & BERT$_{\text{base}}$ & 84.9/85.5 & 91.4 & 92.1 & 93.2 & 59.7 & 90.1 & 86.3 & 72.2  & 83.9 \\ 
2 & BiBERT Baseline & 45.8/47.0 & 73.2 & 66.4 & 77.6 & 11.7 & 7.6 & 70.2 & 54.1 & 50.4 \\ 
3 & BiBERT & 66.1/67.5 & 84.8 & 72.6 & 88.7 & 25.4 & 33.6 & 72.5 & 57.4 & 63.2 \\ 
\hdashline[0.8pt/1pt]
4 & BinaryBERT (Our implementation) & 36.2/35.9 & 59.6 & 52.4 & 65.6 & 9.3 & 19.8 & 69.9 & 52.7 & 45.7 \\
5 & + Our simplied KD  & 37.7/37.3 & 59.5 & 56.8 & 73.4 & 4.1  & 24.8 & 70.8 & 57.0 & 48.0 \\
6 & + Our two-set binarization (Strong Baseline) & \textbf{57.4/59.1} & \textbf{68.3} & \textbf{64.7} & \textbf{81.0} & \textbf{18.2} & \textbf{24.7} & \textbf{71.8} & \textbf{56.7} & \textbf{55.3} \\ 
\hdashline[0.8pt/1pt] 
7 & + Elastic binarization (\ours{} $\ddag$) & 77.1/77.5 & 82.9 & 85.7 & 87.7 & 25.1 & 71.1 & 79.7 & 58.8 & 71.0 \\
8 & + Multi-Distillation (\ours{}) & \textbf{79.5/79.4} & \textbf{85.4} & \textbf{86.4} & \textbf{89.9} & \textbf{32.9} & \textbf{72.0} & \textbf{79.9} & \textbf{62.1} & \textbf{73.5}\\
\hline
\end{tabular}}
\end{table}

\subsection{SQuAD results}
\begin{wraptable}[10]{r}{0.5\textwidth}
\vspace{-1.2em}
\centering
\caption{Comparison of BERT quantization methods on SQuADv1.1 dev set. Metrics are exact match and F1 score.}
\label{tab:squad}
\setlength{\tabcolsep}{2mm}
\resizebox{0.35\textwidth}{!}{
\begin{tabular}{lcc}
\hline
\vspace{0.25pt}
\textbf{Quant} & \begin{tabular}[c]{@{}c@{}}\textbf{\#Bits}\end{tabular} & \textbf{SQuADv1.1}$_\text{EM/F1}$ \\ 
\hline
BERT$_{\text{base}}$ & 32-32-32 & 82.6/89.7 \\ 
\hdashline[0.8pt/1pt]
{BinaryBERT} & 1-1-4 & 77.9/85.8 \\ 
{BinaryBERT} & 1-1-2 & 72.3/81.8 \\
{BinaryBERT} & 1-1-1 & 1.5/8.2 \\
{BiBERT} & 1-1-1 &  8.5/18.9 \\
\hdashline[0.8pt/1pt]
\ours{} & \textbf{1-1-1} & \textbf{63.1/74.9}\\
\hline
\end{tabular}}
\end{wraptable}
We also evaluate on the popular machine reading comprehension (MRC) dataset from \citet{rajpurkar2016squad}.  We compare to our own implementation of the MRC task on top of the BiBERT codebase, since SQuAD results are not reported in that work.  We also show results using the BinaryBERT codebase, without using weight splitting.  The results (Table \ref{tab:squad}) show that this task is significantly harder than most document classification benchmarks in GLUE, and previous binarization methods fail to achieve any meaningful level of performance.  \ours{} does much better, but still trails the 32-bit baseline by 14.8 points in F1.  We conclude that despite the improvements we have demonstrated on the GLUE benchmark, binarizing transformer models accurately is far from a solved problem in general.

\subsection{Ablations}\label{sec:ablations}
We start from the basic binarization implementation from~\citet{bai2021binarybert}, and add each of our contributions in sequence to get a better idea how each contributes to the performance.  The results are shown in Table~\ref{tab:ablation_glue}.

We start by removing attention distillation (Section \ref{sec:minor_improvements}), which results in a 2.3\% improvement (row 5 vs. 4).  Then switching to our two-set binarization (Section \ref{sec:two_set}), which binarizes the attention scores differently than the feed-forward activations,  which gives an additional 7.3\% boost (row 6).  This results in a much stronger baseline than what was used in prior works (row 2).

Moving from fixed to elastic binarization (Section \ref{sec:activation_scaling}) proves hugely important, pushing the average accuracy to 71.0\% (row 7) from only 55.3\% (row 6).  Note that this model already outperforms the current state-of-the-art (row 3) by 7.8\% points.  Finally, we add multi-step distillation (Section \ref{sec:whisky}), which adds another 2.5 points, reaching the final accuracy of 73.5\% on the GLUE benchmark.

\subsection{Learned parameter visualization}\label{sec:visualization}
\begin{wrapfigure}[9]{r}{0.58\textwidth}
    \centering
    \vspace{-1.2em}
    \includegraphics[width=1\linewidth]{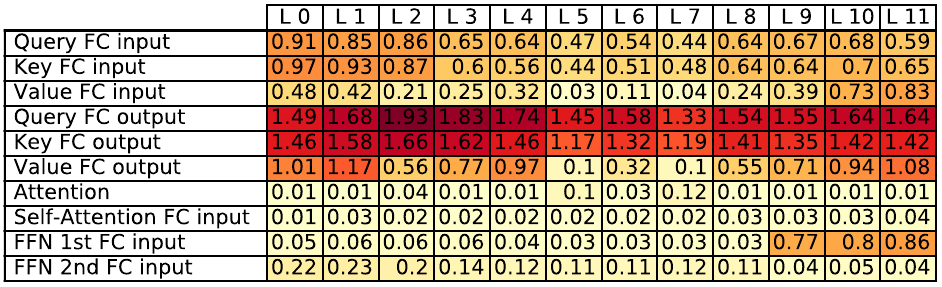}
    \caption{The optimized scaling factor in \ours{}}
    \label{fig:alpha}
\end{wrapfigure}

We visualize the optimized $\alpha$ in the final \ours{} model. As we can see from Figure~\ref{fig:alpha}, the values of the $\alpha$ parameters vary significantly from layer to layer, and have apparent patterns according to layer characteristics. For example, the attention layers need to distribute the attention to different entries, thus the scaling factor for the attentions are learned to be small, while the scaling factors for the query and key outputs are usually larger. Note that the biggest $\alpha$ value is $200\times$ of the smallest $\alpha$, suggesting the importance of learning $\alpha$ dynamically.

\subsection{Exploring multi-distillation paths}\label{sec:paths}

So far we have only considered the fixed quantization schedule, W32A32 $\rightarrow$ W1A2 $\rightarrow$ W1A1.  This is motivated by early experiments showing that one-step distillation to W1A2 works reasonably well.  We explored other optimal schedules, such as distilling to W1A8 resulted in a higher accuracy model, thus a better teacher to distill down to the eventual W1A1 student.  This suggests a trade-off between the quality of the intermediate model, vs. the closeness to the target quantization level.

\begin{wrapfigure}[16]{r}{0.5\textwidth}
    \vspace{-1em}
    \centering
    \includegraphics[width=0.9\linewidth]{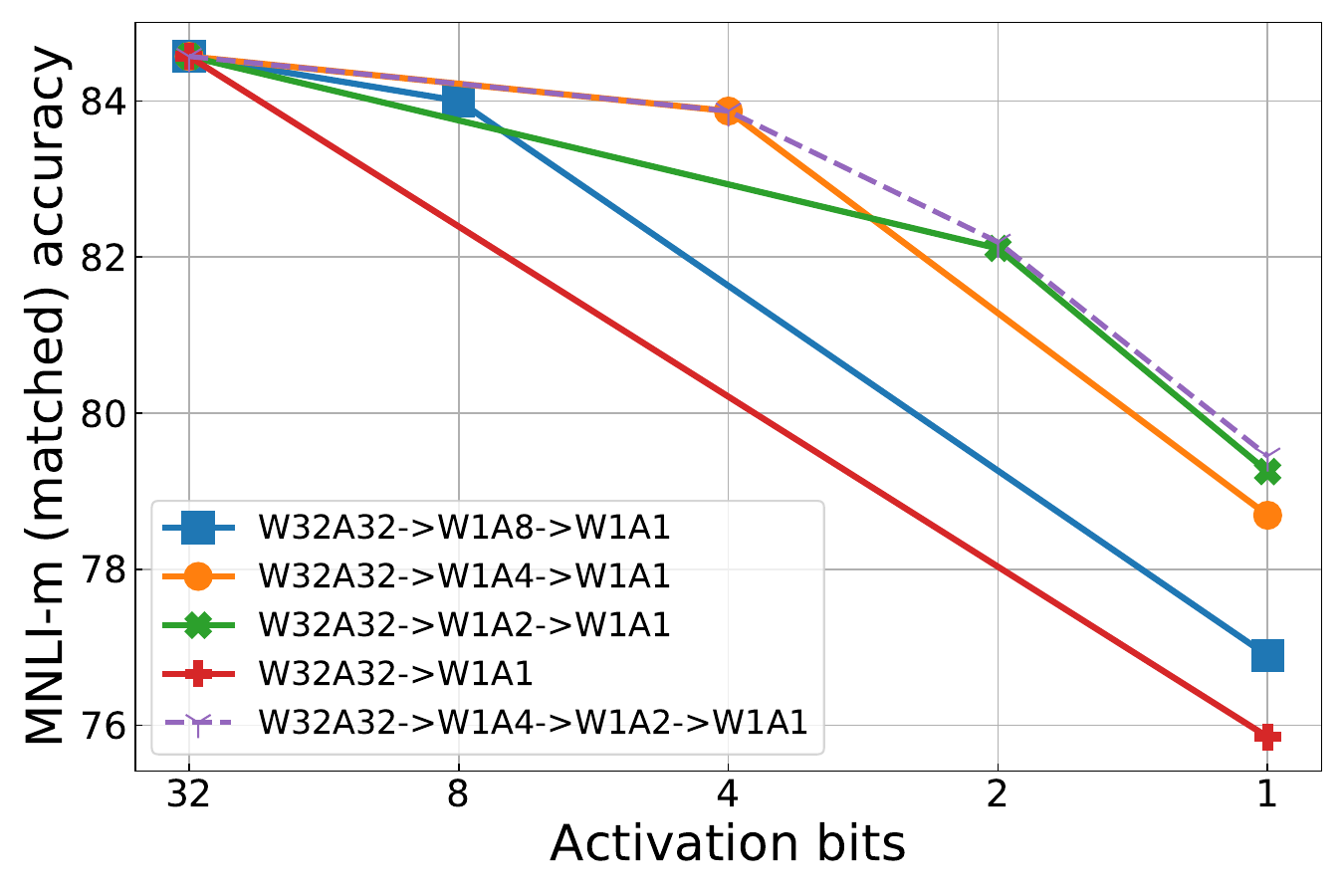}
    \caption{MNLI-m accuracy on various distillation paths.  Each curve represents the sequence of models on a particular quantization schedule.}
    \label{fig:paths}
\end{wrapfigure}

Figure \ref{fig:paths}, illustrates this trade-off. We can see that two-step distillation improves over one-step in every case. While higher precision intermediate models are better as expected, it is better to use a lower precision teacher in the last step since it makes the learning task for the binary student model easier. The closeness to the target quantization is favored despite the lower accuracy of teacher model.

It is of course possible, though more cumbersome, to perform more than two distillation steps.  We also experiment with a three-step schedule, W32A32 $\rightarrow$ W1A4 $\rightarrow$ W1A2 $\rightarrow$ W1A1, which is plotted in the same figure (dashed line).  We find that this particular 3-step schedule does not improve over the 2-step schedule W32A32 $\rightarrow$ W1A2 $\rightarrow$ W1A1.  While this result does not preclude existence of other more optimal schedules, we hypothesize that this is unlikely.

\section{Related work}
\paragraph{Convolutional neural network quantization}
Neural network quantization is a practical tool for compressing model size and reduce storage~\citep{hubara2017quantized}. Quantization for convolutional neural networks has been studied both in the uniform quantization ~\citep{choi2018pact,zhou2016dorefa,gong2019differentiable} and non-uniform quantization~\citep{zhang2018lq,miyashita2016convolutional,li2020additive} settings. The quantization level has been progressively increased, from 8-bit~\citep{wang2018training,zhu2020towards} to 4-bit~\citep{jung2019learning,liu2022nonuniform} and finally to the extreme 1-bit case~\citep{courbariaux2016binarized,rastegari2016xnor,liu2018bi,martinez2020training}.

\paragraph{Transformer quantization} Compared to the CNNs, transformers with attention layers are naturally more challenging to quantize~\citep{bondarenko2021understanding}.  Previous research mainly focused on 8-bit quantization~\citep{zafrir2019q8bert, fan2020training} or 4-bit quantization~\citep{shen2020q,zadeh2020gobo}.
Extremely low-bit quantization for transformers has only been attempted very recently. TernaryBERT~\citep{TernaryBERT} proposed to ternarize the full-precision weights of a fine-tuned BERT model. As a follow-up to TernaryBERT, weight binarization was proposed in~\citet{bai2021binarybert}. Here, the network is trained by first training a ternary half-sized model, which is used as initialization.  Then a weight-splitting step results in a full-sized binarized model, which is further fine-tuned in a subsequent distillation step.  Binarizing both weight and activations in a transformer has proved to be challenging. BiBERT~\citep{qin2021bibert} made the first attempt in this direction with limited success.  Their model performed 20\% worse than a real-valued baseline on the GLUE benchmark~\citep{wang2018glue}, which even underperforms the original LSTM baselines.

\section{Conclusion}
Large pre-trained transformers have transformed NLP and are positioned to serve as the backbone for all AI models.  In this work, we presented the first successful demonstration of a fully binary pre-trained transformer model.  While our approach is general and can be applied to any transformer, we have limited our evaluation to BERT-based models on the GLUE and SQuAD benchmarks.  It remains to be seen how our conclusions will hold when applied to the wide variety of pre-trained transformer models which have gained popularity in recent years, from small mobile models, to gigantic ones with hundreds of billions of parameters.  It will also be interesting to see the performance of the approach on different domains (such as image and speech processing) and tasks (such as text and image generation).  Demonstrating the generality of this approach in a wider setting should significantly widen its impact, therefore we identify this as an important future direction.

\bibliography{main}

\begin{thebibliography}{55}
\providecommand{\natexlab}[1]{#1}
\providecommand{\url}[1]{\texttt{#1}}
\expandafter\ifx\csname urlstyle\endcsname\relax
  \providecommand{\doi}[1]{doi: #1}\else
  \providecommand{\doi}{doi: \begingroup \urlstyle{rm}\Url}\fi

\bibitem[Aguilar et~al.(2020)Aguilar, Ling, Zhang, Yao, Fan, and
  Guo]{aguilar2020knowledge}
Gustavo Aguilar, Yuan Ling, Yu~Zhang, Benjamin Yao, Xing Fan, and Chenlei Guo.
\newblock Knowledge distillation from internal representations.
\newblock In \emph{Proceedings of the AAAI Conference on Artificial
  Intelligence}, volume~34, pp.\  7350--7357, 2020.

\bibitem[Bai et~al.(2021)Bai, Zhang, Hou, Shang, Jin, Jiang, Liu, Lyu, and
  King]{bai2021binarybert}
Haoli Bai, Wei Zhang, Lu~Hou, Lifeng Shang, Jin Jin, Xin Jiang, Qun Liu,
  Michael~R Lyu, and Irwin King.
\newblock Binarybert: Pushing the limit of bert quantization.
\newblock In \emph{ACL/IJCNLP (1)}, 2021.

\bibitem[Bengio et~al.(2013)Bengio, L{\'e}onard, and
  Courville]{bengio2013estimating}
Yoshua Bengio, Nicholas L{\'e}onard, and Aaron Courville.
\newblock Estimating or propagating gradients through stochastic neurons for
  conditional computation.
\newblock \emph{arXiv preprint arXiv:1308.3432}, 2013.

\bibitem[Bentivogli et~al.(2009)Bentivogli, Clark, Dagan, and
  Giampiccolo]{bentivogli2009fifth}
Luisa Bentivogli, Peter Clark, Ido Dagan, and Danilo Giampiccolo.
\newblock The fifth pascal recognizing textual entailment challenge.
\newblock In \emph{TAC}, 2009.

\bibitem[Bondarenko et~al.(2021)Bondarenko, Nagel, and
  Blankevoort]{bondarenko2021understanding}
Yelysei Bondarenko, Markus Nagel, and Tijmen Blankevoort.
\newblock Understanding and overcoming the challenges of efficient transformer
  quantization.
\newblock In \emph{Proceedings of the 2021 Conference on Empirical Methods in
  Natural Language Processing}, pp.\  7947--7969, 2021.

\bibitem[Bridle(1989)]{bridle1989training}
John Bridle.
\newblock Training stochastic model recognition algorithms as networks can lead
  to maximum mutual information estimation of parameters.
\newblock \emph{Advances in neural information processing systems}, 2, 1989.

\bibitem[Brown et~al.(2020)Brown, Mann, Ryder, Subbiah, Kaplan, Dhariwal,
  Neelakantan, Shyam, Sastry, Askell, et~al.]{brown2020language}
Tom Brown, Benjamin Mann, Nick Ryder, Melanie Subbiah, Jared~D Kaplan, Prafulla
  Dhariwal, Arvind Neelakantan, Pranav Shyam, Girish Sastry, Amanda Askell,
  et~al.
\newblock Language models are few-shot learners.
\newblock \emph{Advances in neural information processing systems},
  33:\penalty0 1877--1901, 2020.

\bibitem[Cer et~al.(2017)Cer, Diab, Agirre, Lopez-Gazpio, and
  Specia]{cer2017semeval}
Daniel Cer, Mona Diab, Eneko Agirre, Inigo Lopez-Gazpio, and Lucia Specia.
\newblock Semeval-2017 task 1: Semantic textual similarity-multilingual and
  cross-lingual focused evaluation.
\newblock \emph{arXiv preprint arXiv:1708.00055}, 2017.

\bibitem[Chen et~al.(2018)Chen, Zhang, Zhang, and Zhao]{chen2018quora}
Zihan Chen, Hongbo Zhang, Xiaoji Zhang, and Leqi Zhao.
\newblock Quora question pairs.
\newblock \emph{University of Waterloo}, pp.\  1--7, 2018.

\bibitem[Choi et~al.(2018)Choi, Wang, Venkataramani, et~al.]{choi2018pact}
Jungwook Choi, Zhuo Wang, Swagath Venkataramani, et~al.
\newblock Pact: Parameterized clipping activation for quantized neural
  networks.
\newblock \emph{arXiv e-prints}, pp.\  arXiv--1805, 2018.

\bibitem[Courbariaux et~al.(2016)Courbariaux, Hubara, Soudry, El-Yaniv, and
  Bengio]{courbariaux2016binarized}
Matthieu Courbariaux, Itay Hubara, Daniel Soudry, Ran El-Yaniv, and Yoshua
  Bengio.
\newblock Binarized neural networks: Training deep neural networks with weights
  and activations constrained to+ 1 or-1.
\newblock \emph{arXiv preprint arXiv:1602.02830}, 2016.

\bibitem[Devlin et~al.(2019)Devlin, Chang, Lee, and Toutanova]{devlin2019bert}
Jacob Devlin, Ming-Wei Chang, Kenton Lee, and Kristina Toutanova.
\newblock Bert: Pre-training of deep bidirectional transformers for language
  understanding.
\newblock In \emph{NAACL-HLT (1)}, 2019.

\bibitem[Dolan \& Brockett(2005)Dolan and Brockett]{dolan2005automatically}
Bill Dolan and Chris Brockett.
\newblock Automatically constructing a corpus of sentential paraphrases.
\newblock In \emph{Third International Workshop on Paraphrasing (IWP2005)},
  2005.

\bibitem[Esser et~al.(2019)Esser, McKinstry, Bablani, Appuswamy, and
  Modha]{esser2019learned}
Steven~K Esser, Jeffrey~L McKinstry, Deepika Bablani, Rathinakumar Appuswamy,
  and Dharmendra~S Modha.
\newblock Learned step size quantization.
\newblock In \emph{International Conference on Learning Representations}, 2019.

\bibitem[Fan et~al.(2020)Fan, Stock, Graham, Grave, Gribonval, Jegou, and
  Joulin]{fan2020training}
Angela Fan, Pierre Stock, Benjamin Graham, Edouard Grave, R{\'e}mi Gribonval,
  Herve Jegou, and Armand Joulin.
\newblock Training with quantization noise for extreme model compression.
\newblock \emph{arXiv preprint arXiv:2004.07320}, 2020.

\bibitem[Gholami et~al.(2021)Gholami, Kim, Dong, Yao, Mahoney, and
  Keutzer]{gholami2021survey}
Amir Gholami, Sehoon Kim, Zhen Dong, Zhewei Yao, Michael~W Mahoney, and Kurt
  Keutzer.
\newblock A survey of quantization methods for efficient neural network
  inference.
\newblock \emph{arXiv preprint arXiv:2103.13630}, 2021.

\bibitem[Gong et~al.(2019)Gong, Liu, Jiang, Li, Hu, Lin, Yu, and
  Yan]{gong2019differentiable}
Ruihao Gong, Xianglong Liu, Shenghu Jiang, Tianxiang Li, Peng Hu, Jiazhen Lin,
  Fengwei Yu, and Junjie Yan.
\newblock Differentiable soft quantization: Bridging full-precision and low-bit
  neural networks.
\newblock In \emph{Proceedings of the IEEE/CVF International Conference on
  Computer Vision}, pp.\  4852--4861, 2019.

\bibitem[Han \& Moraga(1995)Han and Moraga]{han1995influence}
Jun Han and Claudio Moraga.
\newblock The influence of the sigmoid function parameters on the speed of
  backpropagation learning.
\newblock In \emph{International workshop on artificial neural networks}, pp.\
  195--201. Springer, 1995.

\bibitem[Hendrycks \& Gimpel(2016)Hendrycks and Gimpel]{hendrycks2016gaussian}
Dan Hendrycks and Kevin Gimpel.
\newblock Gaussian error linear units (gelus).
\newblock \emph{arXiv preprint arXiv:1606.08415}, 2016.

\bibitem[Hinton et~al.(2015)Hinton, Vinyals, Dean,
  et~al.]{hinton2015distilling}
Geoffrey Hinton, Oriol Vinyals, Jeff Dean, et~al.
\newblock Distilling the knowledge in a neural network.
\newblock \emph{arXiv preprint arXiv:1503.02531}, 2\penalty0 (7), 2015.

\bibitem[Hubara et~al.(2017)Hubara, Courbariaux, Soudry, El-Yaniv, and
  Bengio]{hubara2017quantized}
Itay Hubara, Matthieu Courbariaux, Daniel Soudry, Ran El-Yaniv, and Yoshua
  Bengio.
\newblock Quantized neural networks: Training neural networks with low
  precision weights and activations.
\newblock \emph{The Journal of Machine Learning Research}, 18\penalty0
  (1):\penalty0 6869--6898, 2017.

\bibitem[Jung et~al.(2019)Jung, Son, Lee, Son, Han, Kwak, Hwang, and
  Choi]{jung2019learning}
Sangil Jung, Changyong Son, Seohyung Lee, Jinwoo Son, Jae-Joon Han, Youngjun
  Kwak, Sung~Ju Hwang, and Changkyu Choi.
\newblock Learning to quantize deep networks by optimizing quantization
  intervals with task loss.
\newblock In \emph{Proceedings of the IEEE/CVF Conference on Computer Vision
  and Pattern Recognition}, pp.\  4350--4359, 2019.

\bibitem[Kim et~al.(2019)Kim, Bhalgat, Lee, Patel, and Kwak]{kim2019qkd}
Jangho Kim, Yash Bhalgat, Jinwon Lee, Chirag Patel, and Nojun Kwak.
\newblock Qkd: Quantization-aware knowledge distillation.
\newblock \emph{arXiv preprint arXiv:1911.12491}, 2019.

\bibitem[Li et~al.(2020)Li, Dong, and Wang]{li2020additive}
Yuhang Li, Xin Dong, and Wei Wang.
\newblock Additive powers-of-two quantization: An efficient non-uniform
  discretization for neural networks.
\newblock In \emph{International Conference on Learning Representations}, 2020.

\bibitem[Liu et~al.(2019)Liu, Ott, Goyal, Du, Joshi, Chen, Levy, Lewis,
  Zettlemoyer, and Stoyanov]{liu2019roberta}
Yinhan Liu, Myle Ott, Naman Goyal, Jingfei Du, Mandar Joshi, Danqi Chen, Omer
  Levy, Mike Lewis, Luke Zettlemoyer, and Veselin Stoyanov.
\newblock Roberta: A robustly optimized bert pretraining approach.
\newblock \emph{arXiv preprint arXiv:1907.11692}, 2019.

\bibitem[Liu et~al.(2018)Liu, Wu, Luo, Yang, Liu, and Cheng]{liu2018bi}
Zechun Liu, Baoyuan Wu, Wenhan Luo, Xin Yang, Wei Liu, and Kwang-Ting Cheng.
\newblock Bi-real net: Enhancing the performance of 1-bit cnns with improved
  representational capability and advanced training algorithm.
\newblock In \emph{Proceedings of the European conference on computer vision
  (ECCV)}, pp.\  722--737, 2018.

\bibitem[Liu et~al.(2020)Liu, Shen, Savvides, and Cheng]{liu2020reactnet}
Zechun Liu, Zhiqiang Shen, Marios Savvides, and Kwang-Ting Cheng.
\newblock Reactnet: Towards precise binary neural network with generalized
  activation functions.
\newblock In \emph{European Conference on Computer Vision}, pp.\  143--159.
  Springer, 2020.

\bibitem[Liu et~al.(2022)Liu, Cheng, Huang, Xing, and Shen]{liu2022nonuniform}
Zechun Liu, Kwang-Ting Cheng, Dong Huang, Eric~P Xing, and Zhiqiang Shen.
\newblock Nonuniform-to-uniform quantization: Towards accurate quantization via
  generalized straight-through estimation.
\newblock In \emph{Proceedings of the IEEE/CVF Conference on Computer Vision
  and Pattern Recognition}, pp.\  4942--4952, 2022.

\bibitem[Martinez et~al.(2020)Martinez, Yang, Bulat, and
  Tzimiropoulos]{martinez2020training}
Brais Martinez, Jing Yang, Adrian Bulat, and Georgios Tzimiropoulos.
\newblock Training binary neural networks with real-to-binary convolutions.
\newblock In \emph{ICLR}, 2020.

\bibitem[Miyashita et~al.(2016)Miyashita, Lee, and
  Murmann]{miyashita2016convolutional}
Daisuke Miyashita, Edward~H Lee, and Boris Murmann.
\newblock Convolutional neural networks using logarithmic data representation.
\newblock \emph{arXiv preprint arXiv:1603.01025}, 2016.

\bibitem[Nair \& Hinton(2010)Nair and Hinton]{nair2010rectified}
Vinod Nair and Geoffrey~E Hinton.
\newblock Rectified linear units improve restricted boltzmann machines.
\newblock In \emph{Icml}, 2010.

\bibitem[Nurvitadhi et~al.(2016)Nurvitadhi, Sheffield, Sim, Mishra, Venkatesh,
  and Marr]{nurvitadhi2016accelerating}
Eriko Nurvitadhi, David Sheffield, Jaewoong Sim, Asit Mishra, Ganesh Venkatesh,
  and Debbie Marr.
\newblock Accelerating binarized neural networks: Comparison of fpga, cpu, gpu,
  and asic.
\newblock In \emph{2016 International Conference on Field-Programmable
  Technology (FPT)}, pp.\  77--84. IEEE, 2016.

\bibitem[Qin et~al.(2020)Qin, Gong, Liu, Shen, Wei, Yu, and Song]{IR-Net}
Haotong Qin, Ruihao Gong, Xianglong Liu, Mingzhu Shen, Ziran Wei, Fengwei Yu,
  and Jingkuan Song.
\newblock Forward and backward information retention for accurate binary neural
  networks.
\newblock In \emph{CVPR}, 2020.

\bibitem[Qin et~al.(2021)Qin, Ding, Zhang, Qinghua, Liu, Dang, Liu, and
  Liu]{qin2021bibert}
Haotong Qin, Yifu Ding, Mingyuan Zhang, YAN Qinghua, Aishan Liu, Qingqing Dang,
  Ziwei Liu, and Xianglong Liu.
\newblock Bibert: Accurate fully binarized bert.
\newblock In \emph{International Conference on Learning Representations}, 2021.

\bibitem[Radford et~al.(2018)Radford, Narasimhan, Salimans, and
  Sutskever]{radford2018improving}
Alec Radford, Karthik Narasimhan, Tim Salimans, and Ilya Sutskever.
\newblock Improving language understanding by generative pre-training.
\newblock 2018.

\bibitem[Radford et~al.(2019)Radford, Wu, Child, Luan, Amodei, Sutskever,
  et~al.]{radford2019language}
Alec Radford, Jeffrey Wu, Rewon Child, David Luan, Dario Amodei, Ilya
  Sutskever, et~al.
\newblock Language models are unsupervised multitask learners.
\newblock \emph{OpenAI blog}, 1\penalty0 (8):\penalty0 9, 2019.

\bibitem[Raffel et~al.(2020)Raffel, Shazeer, Roberts, Lee, Narang, Matena,
  Zhou, Li, and Liu]{raffel2020exploring}
Colin Raffel, Noam Shazeer, Adam Roberts, Katherine Lee, Sharan Narang, Michael
  Matena, Yanqi Zhou, Wei Li, and Peter~J Liu.
\newblock Exploring the limits of transfer learning with a unified text-to-text
  transformer.
\newblock \emph{Journal of Machine Learning Research}, 21:\penalty0 1--67,
  2020.

\bibitem[Rajpurkar et~al.(2016)Rajpurkar, Zhang, Lopyrev, and
  Liang]{rajpurkar2016squad}
Pranav Rajpurkar, Jian Zhang, Konstantin Lopyrev, and Percy Liang.
\newblock Squad: 100, 000+ questions for machine comprehension of text.
\newblock In \emph{EMNLP}, 2016.

\bibitem[Rastegari et~al.(2016)Rastegari, Ordonez, Redmon, and
  Farhadi]{rastegari2016xnor}
Mohammad Rastegari, Vicente Ordonez, Joseph Redmon, and Ali Farhadi.
\newblock Xnor-net: Imagenet classification using binary convolutional neural
  networks.
\newblock In \emph{European conference on computer vision}, pp.\  525--542.
  Springer, 2016.

\bibitem[Sanh et~al.(2019)Sanh, Debut, Chaumond, and Wolf]{sanh2019distilbert}
Victor Sanh, Lysandre Debut, Julien Chaumond, and Thomas Wolf.
\newblock Distilbert, a distilled version of bert: smaller, faster, cheaper and
  lighter.
\newblock \emph{arXiv preprint arXiv:1910.01108}, 2019.

\bibitem[Shen et~al.(2020)Shen, Dong, Ye, Ma, Yao, Gholami, Mahoney, and
  Keutzer]{shen2020q}
Sheng Shen, Zhen Dong, Jiayu Ye, Linjian Ma, Zhewei Yao, Amir Gholami,
  Michael~W Mahoney, and Kurt Keutzer.
\newblock Q-bert: Hessian based ultra low precision quantization of bert.
\newblock In \emph{Proceedings of the AAAI Conference on Artificial
  Intelligence}, volume~34, pp.\  8815--8821, 2020.

\bibitem[Socher et~al.(2013)Socher, Perelygin, Wu, Chuang, Manning, Ng, and
  Potts]{socher2013recursive}
Richard Socher, Alex Perelygin, Jean Wu, Jason Chuang, Christopher~D Manning,
  Andrew~Y Ng, and Christopher Potts.
\newblock Recursive deep models for semantic compositionality over a sentiment
  treebank.
\newblock In \emph{Proceedings of the 2013 conference on empirical methods in
  natural language processing}, pp.\  1631--1642, 2013.

\bibitem[Vaswani et~al.(2017)Vaswani, Shazeer, Parmar, Uszkoreit, Jones, Gomez,
  Kaiser, and Polosukhin]{vaswani2017attention}
Ashish Vaswani, Noam Shazeer, Niki Parmar, Jakob Uszkoreit, Llion Jones,
  Aidan~N Gomez, {\L}ukasz Kaiser, and Illia Polosukhin.
\newblock Attention is all you need.
\newblock In \emph{NeurIPS}, 2017.

\bibitem[Wang et~al.(2019)Wang, Singh, Michael, Hill, Levy, and
  Bowman]{wang2018glue}
Alex Wang, Amanpreet Singh, Julian Michael, Felix Hill, Omer Levy, and Samuel~R
  Bowman.
\newblock Glue: A multi-task benchmark and analysis platform for natural
  language understanding.
\newblock In \emph{International Conference on Learning Representations}, 2019.

\bibitem[Wang et~al.(2018)Wang, Choi, Brand, Chen, and
  Gopalakrishnan]{wang2018training}
Naigang Wang, Jungwook Choi, Daniel Brand, Chia-Yu Chen, and Kailash
  Gopalakrishnan.
\newblock Training deep neural networks with 8-bit floating point numbers.
\newblock \emph{Advances in neural information processing systems}, 31, 2018.

\bibitem[Warstadt et~al.(2019)Warstadt, Singh, and Bowman]{warstadt2019cola}
Alex Warstadt, Amanpreet Singh, and Samuel~R Bowman.
\newblock Cola: The corpus of linguistic acceptability (with added
  annotations).
\newblock 2019.

\bibitem[Williams et~al.(2018)Williams, Nangia, and Bowman]{williams2018multi}
Adina Williams, Nikita Nangia, and Samuel~R Bowman.
\newblock The multi-genre nli corpus.
\newblock 2018.

\bibitem[Yang et~al.(2019)Yang, Huang, Wu, Zhang, Ma, Gambardella, Blott,
  Lavagno, Vissers, Wawrzynek, et~al.]{yang2019synetgy}
Yifan Yang, Qijing Huang, Bichen Wu, Tianjun Zhang, Liang Ma, Giulio
  Gambardella, Michaela Blott, Luciano Lavagno, Kees Vissers, John Wawrzynek,
  et~al.
\newblock Synetgy: Algorithm-hardware co-design for convnet accelerators on
  embedded fpgas.
\newblock In \emph{Proceedings of the 2019 ACM/SIGDA international symposium on
  field-programmable gate arrays}, pp.\  23--32, 2019.

\bibitem[Zadeh et~al.(2020)Zadeh, Edo, Awad, and Moshovos]{zadeh2020gobo}
Ali~Hadi Zadeh, Isak Edo, Omar~Mohamed Awad, and Andreas Moshovos.
\newblock Gobo: Quantizing attention-based nlp models for low latency and
  energy efficient inference.
\newblock In \emph{2020 53rd Annual IEEE/ACM International Symposium on
  Microarchitecture (MICRO)}, pp.\  811--824. IEEE, 2020.

\bibitem[Zafrir et~al.(2019)Zafrir, Boudoukh, Izsak, and
  Wasserblat]{zafrir2019q8bert}
Ofir Zafrir, Guy Boudoukh, Peter Izsak, and Moshe Wasserblat.
\newblock Q8bert: Quantized 8bit bert.
\newblock In \emph{2019 Fifth Workshop on Energy Efficient Machine Learning and
  Cognitive Computing-NeurIPS Edition (EMC2-NIPS)}, pp.\  36--39. IEEE, 2019.

\bibitem[Zhang et~al.(2018)Zhang, Yang, Ye, and Hua]{zhang2018lq}
Dongqing Zhang, Jiaolong Yang, Dongqiangzi Ye, and Gang Hua.
\newblock Lq-nets: Learned quantization for highly accurate and compact deep
  neural networks.
\newblock In \emph{Proceedings of the European conference on computer vision
  (ECCV)}, pp.\  365--382, 2018.

\bibitem[Zhang et~al.(2020)Zhang, Hou, Yin, Shang, Chen, Jiang, and
  Liu]{TernaryBERT}
Wei Zhang, Lu~Hou, Yichun Yin, Lifeng Shang, Xiao Chen, Xin Jiang, and Qun Liu.
\newblock Ternarybert: Distillation-aware ultra-low bit {BERT}.
\newblock In Bonnie Webber, Trevor Cohn, Yulan He, and Yang Liu (eds.),
  \emph{EMNLP}, 2020.

\bibitem[Zhou et~al.(2016)Zhou, Wu, Ni, Zhou, Wen, and Zou]{zhou2016dorefa}
Shuchang Zhou, Yuxin Wu, Zekun Ni, Xinyu Zhou, He~Wen, and Yuheng Zou.
\newblock Dorefa-net: Training low bitwidth convolutional neural networks with
  low bitwidth gradients.
\newblock \emph{arXiv preprint arXiv:1606.06160}, 2016.

\bibitem[Zhu et~al.(2020)Zhu, Gong, Yu, Liu, Wang, Li, Yang, and
  Yan]{zhu2020towards}
Feng Zhu, Ruihao Gong, Fengwei Yu, Xianglong Liu, Yanfei Wang, Zhelong Li,
  Xiuqi Yang, and Junjie Yan.
\newblock Towards unified int8 training for convolutional neural network.
\newblock In \emph{Proceedings of the IEEE/CVF Conference on Computer Vision
  and Pattern Recognition}, pp.\  1969--1979, 2020.

\bibitem[Zhuang et~al.(2018)Zhuang, Shen, Tan, Liu, and
  Reid]{zhuang2018towards}
Bohan Zhuang, Chunhua Shen, Mingkui Tan, Lingqiao Liu, and Ian Reid.
\newblock Towards effective low-bitwidth convolutional neural networks.
\newblock In \emph{Proceedings of the IEEE conference on computer vision and
  pattern recognition}, pp.\  7920--7928, 2018.

\end{thebibliography}
\bibliographystyle{main}

\appendix
\section{Appendix}

\subsection{\ours{} vs. progressive distillation}\label{sec:progressive_ablation}
\begin{table}[h]
\centering
\caption{\ours{} vs. progressive distillation on selected GLUE tasks. Methods differ in the teacher model used and the model from which the student weights are initialized.}
\label{tab:progressive}
\setlength{\tabcolsep}{1.4mm}
\resizebox{0.8\textwidth}{!}{
\begin{tabular}{lcccccccccccc}
\hline
\vspace{0.25pt}
\textbf{Method} & \textbf{Teacher} & \textbf{Initialization} & \textbf{MNLI}$_\text{-m/mm}$ & \textbf{QQP} & \textbf{QNLI} & \textbf{SST-2} & \textbf{CoLA} & \textbf{STS-B} & \textbf{MRPC} & \textbf{RTE} & \textbf{Avg.}\\ 
\hline
{BiBERT Distillation} & 32-32-32 & 32-32-32 & 77.0/77.2 & 83.1 & 84.1 & 89.7 & 31.3 & 60.1 & 75.5 & 56.7 & 69.7 \\ 
{Progressive} & 32-32-32 & 1-1-2 & 78.9/78.9 & 85.0 & 86.4 & 89.6 & 30.5 & \textbf{75.1} & \textbf{81.1} & 60.6 & 73.4 \\
\ours{} & \textbf{1-1-2} & \textbf{1-1-2} & \textbf{79.5/79.4} & \textbf{85.4} & \textbf{86.4} & \textbf{89.9} & \textbf{32.9} & 72.0 & 79.9 & \textbf{62.1} & \textbf{73.5} \\
\hline
\end{tabular}}
\end{table}

Previous work has also recognized the importance of good initialization for binary model training, and proposed to perform distillation while progressively quantizing the student model \citep{zhuang2018towards, yang2019synetgy}.  Progressive distillation ensures a good initialization for the student model at each step.  However, in this approach the teacher model is fixed to the full precision model, which does not address the problem of teacher-student gap.  In Table \ref{tab:progressive} we compare \ours{} to a comparable implementation of progressive distillation, using the same quantization schedule, W32A32 $\rightarrow$ W1A2 $\rightarrow$ W1A1, as ours.  We keep the teacher model fixed, while re-initializing the student model from the latest quantized version at each step.  We see that using a quantized teacher model is helpful, especially in the high-data regime. However, our method can lag behind progressive distillation for small datasets such as STS-B and MRPC.

\subsection{Elastic binarization function vs. ReActNet learnable bias}
\label{sec:vs_reactnet}
\begin{table}[h]
\centering
\caption{Elastic binarization function vs. ReActNet~\citep{liu2020reactnet} learnable bias on GLUE tasks.}
\label{tab:vs_reactnet}
\setlength{\tabcolsep}{1.4mm}
\resizebox{0.8\textwidth}{!}{
\begin{tabular}{lcccccccccccc}
\hline
\vspace{0.25pt}
\textbf{Method} & \textbf{MNLI}$_\text{-m/mm}$ & \textbf{QQP} & \textbf{QNLI} & \textbf{SST-2} & \textbf{CoLA} & \textbf{STS-B} & \textbf{MRPC} & \textbf{RTE} & \textbf{Avg.}\\ 
\hline
Our two-set binarization (Strong Baseline)& 57.4/59.1 & 68.3 & 64.7 & 81.0 & 18.2 & 24.7 & 71.8 & 56.7 & 55.3  \\ 
+ learnable scale & 76.5/76.8 & 82.7 & 85.1 & 88.1 & 26.6 & 62.3 & 74.3 & 58.1 & 69.2 \\
+ learnable scale and bias (BiT $\ddagger$)  & 77.1/77.5 & 82.9 & 85.7 & 87.7 & 25.1 & 71.1 & 79.7 & 58.8 & 71.0 \\
\hline
\end{tabular}}
\end{table}
Inspired by the learnable bias proposed in ReActNet~\citep{liu2020reactnet}, we further propose elastic binarization function to learn both learnable scaling factors and learnable bias. We find this learnable scaling factor critical for the final performance. As shown in table~\ref{tab:vs_reactnet}, the proposed learnable scaling factor brings 13.9\% accuracy improvement, and further adding learnable bias boosts the accuracy by 1.8\%.

\subsection{Two-set binarization scheme vs. Bi-Attention}\label{sec:vs_biattention}
\begin{table}[h]
\centering
\caption{Two-set binarization scheme vs. Bi-Attention~\citep{qin2021bibert} on GLUE tasks. Methods differ in whether using $\rm{SoftMax}$ in attention and whether binarizing the $\rm{ReLU}$ output to \{0 ,1\}.}
\label{tab:vs_biattention}
\setlength{\tabcolsep}{1.4mm}
\resizebox{\textwidth}{!}{
\begin{tabular}{lcccccccccccc}
\hline
\vspace{0.25pt}
\textbf{Method} & \textbf{Attention} & \textbf{ReLU output} & \textbf{MNLI}$_\text{-m/mm}$ & \textbf{QQP} & \textbf{QNLI} & \textbf{SST-2} & \textbf{CoLA} & \textbf{STS-B} & \textbf{MRPC} & \textbf{RTE} & \textbf{Avg.}\\ 
\hline
Bi-Attention (w/o Softmax)& \{0, 1\} & \{-1, 1\}  & 48.1/50.0 & 60.1 & 60.6 & 78.8 & 14.0 & 22.3 & 68.4 & 58.1 & 51.3 \\
Binarize attention to \{0, 1\} (w/ Softmax) & \{0, 1\} & \{-1, 1\} & 51.9/52.6 & 76.2 & 60.5 & 79.6 & 11.6 & 18.1 & 70.6 & 55.6 & 53.0 \\
Two-set binarization scheme & \{0, 1\} & \{0, 1\}& 57.4/59.1 & 68.3 & 64.7 & 81.0 & 18.2 & 24.7 & 71.8 & 56.7 & 55.3 \\

\hline
\end{tabular}}
\end{table}
In contrast to Bi-Attention proposed in BiBERT~\citep{qin2021bibert} that removes $\rm{SoftMax}$ and binarizes the attention to \{0, 1\} with $\rm{bool}$ function, our two-set binarization scheme finds that keeping $\rm{SoftMax}$ in attention computation and also binarizing the positive output of $\rm{ReLU}$ layer to \{0, 1\} works better. We conduct meticulous experiments to compare these choices. In Table~\ref{tab:vs_biattention}, we show that, compared to removing $\rm{SoftMax}$ as Bi-Attention suggested, simply binarizing the activations after $\rm{SoftMax}$ layer to \{0, 1\} even produces 1.7\% better accuracy. Furthermore, binarizing the $\rm{ReLU}$ layer output to \{0, 1\} instead of \{-1, 1\} helps the binary network match real-valued distributions and further brings 2.3\% accuracy improvement.

\subsection{Binary convolution implementation for two-set binarization scheme}
The binary convolution between the weights and activations that are both binarized to \{-1, 1\} (i.e.  $\mathbf{A_B} \in $ \{-1, 1\},  $\mathbf{W_B} \in $ \{-1, 1\}) can be implemented by the bitwise $\rm{xnor}$ operation followed by a $\rm{popcnt}$ operation~\citep{rastegari2016xnor,liu2018bi}:
\begin{equation}
\label{eq:binconv}
    \mathbf{A_B}\cdot\mathbf{W_B} = \rm{popcnt}(\rm{xnor}(\mathbf{A_B},\mathbf{W_B}))
\end{equation}
For the case where activations are binarized to \{0, 1\} in two-set binarization scheme, the binary activation $\mathbf{A_B} \in $ \{0, 1\} can be represented with $\mathbf{A'_B} \in $ \{-1, 1\} through a simple linear mapping:
$ \mathbf{A_B} = \frac{\mathbf{A'_B}+ 1}{2}$.
Thus the matrix computation between binary weights ($\mathbf{W_B}\in $ \{-1, 1\} ) and binary activations ($\mathbf{A_B} \in $ \{0, 1\}) can be converted to the operations between $\mathbf{W_B}\in $ \{-1, 1\} and $\mathbf{A'_B}\in $ \{-1, 1\} as:
\begin{equation}
\mathbf{A_B}\cdot\mathbf{W_B} = (\frac{\mathbf{A'_B} + \mathbf{1}}{2})\cdot\mathbf{W_B}= \frac{1}{2} (\rm{popcnt}(\rm{xnor}(\mathbf{A'_B},\mathbf{W_B}))+ \sum_i\mathbf{W_{B_i}}) 
\end{equation}
Here the $\sum_i\mathbf{W_{B_i}}$ is summing up the values in $\mathbf{W_B}$, which can be pre-computed and stored as bias. Thus in the two-set binarization scheme where activations are binarized to \{0, 1\}, the binary convolution can still be implemented with the general binary convolution in E.q.~\ref{eq:binconv} at no additional complexity cost. 

\subsection{Evaluation benchmarks}\label{sec:glue}

\subsubsection{GLUE}
The GLUE benchmark~\citep{wang2018glue} includes the following datasets:

\paragraph{MNLI} Multi-Genre Natural Language Inference is an entailment classification task~\citep{williams2018multi}.  The goal is to predict whether a given sentence {\it entails}, {\it contradicts}, or is {\it neutral} with respect to another.

\paragraph{QQP} Quora Question Pairs is a paraphrase detection task.  The goal is to classify whether two given questions have the same meaning.  The questions were sourced from the Quora question answering website~\citep{chen2018quora}.

\paragraph{QNLI} Question Natural Language Inference~\citep{wang2018glue} is a binary classification task which is derived from the Stanford Question Answering Dataset~\citep{rajpurkar2016squad}.  The task is to predict whether a sentence contains the answer to a given question.

\paragraph{SST-2} The Stanford Sentiment Treebank is a binary sentiment classification task, with content taken from movie reviews~\citep{socher2013recursive}.

\paragraph{CoLA} The Corpus of Linguistic Acceptability is a corpus of English sentences, each with a binary label denoting whether the sentence is linguistically acceptable~\citep{warstadt2019cola}.

\paragraph{STS-B} The Semantic Textual Similarity Benchmark is a sentence pair classification task.  The goal is to predict how similar the two sentences are in meaning, with scores ranging from 1 to 5~\citep{cer2017semeval}. 

\paragraph{MRPC} Microsoft Research Paraphrase Corpus is another sentence pair paraphrase detection task similar to QQP.  The sentence pairs are sourced from online news sources~\citep{dolan2005automatically}.

\paragraph{RTE} Recognizing Textual Entailment is a small natural language inference dataset similar to MNLI in content~\citep{bentivogli2009fifth}.

\subsubsection{SQuAD}
The SQuAD benchmark~\citep{rajpurkar2016squad}, \textit{i.e.}, Stanford Question Answering Dataset, is a reading comprehension dataset, consisting of questions on a set of Wikipedia articles, where the answer to each question is a segment of text from the corresponding passage, or the question might be unanswerable. 

\subsection{Technical details}\label{sec:technical}
For each experiment, we sweep the learning rate in \{1e-4, 2e-4, 5e-4\} and the batch size in \{8, 16\} for QNLI, SST-2, CoLA, STS-B, MRPC, RTE, and \{16, 32\} for MNLI, QQP as well as SQuAD, and choose the settings with the highest accuracy on the validation set.
We use the same number of training epochs as BiBERT~\citep{qin2021bibert}, \textit{i.e.}, 50 for CoLA, 20 for MRPC, STS-B and RTE, 10 for SST-2 and QNLI, 5 for MNLI and QQP. 
We adopt the Adam optimizer with weight decay 0.01 and use 0.1 warmup ratio with linear learning rate decay. 

Our full precision checkpoints are taken from \url{https://textattack.readthedocs.io/en/latest/3recipes/models.html#bert-base-uncased}.

\end{document}